\documentclass[utf8,usenames,dvipsnames,table,xcdraw]{article} % for Science, Engineering and Humanities and Social Sciences articles

\usepackage[final]{nips_2016}

\usepackage[utf8]{inputenc} % allow utf-8 input
\usepackage[T1]{fontenc}    % use 8-bit T1 fonts
\usepackage{hyperref}       % hyperlinks
\usepackage{url}            % simple URL typesetting
\usepackage{booktabs}       % professional-quality tables
\usepackage{amsfonts}       % blackboard math symbols
\usepackage{nicefrac}       % compact symbols for 1/2, etc.
\usepackage{microtype}      % microtypography
\usepackage{amsfonts}                   % AMS Math Packet (Fonts)
\usepackage{amsmath}                    % AMS Math Packet
\usepackage{graphicx}                   % Inclusion of graphics
\usepackage[usenames,dvipsnames]{xcolor}
\usepackage[labelsep=quad,indention=10pt]{caption}
\usepackage[labelfont=bf,list=true]{subcaption}
\usepackage[onehalfspacing]{setspace}
\usepackage{algorithm}
\usepackage{algorithmic}
\title{Continual Learning from Synthetic Data for a Humanoid Exercise Robot}

\author{
	Nicolas Duczek\,$^{1*}$, Matthias Kerzel\,$^{1}$ and Stefan Wermter\,$^{1}$\\
	$^1$\small{Knowledge Technology, Department of Informatics, University of Hamburg, Hamburg, Germany}\\
}

\begin{document}

\maketitle

\begin{abstract}
\noindent
%%% Leave the Abstract empty if your article does not require one, please see the Summary Table for full details.
In order to detect and correct physical exercises, a Grow-When-Required Network (GWR) with recurrent connections, episodic memory and a novel subnode mechanism is developed in order to learn spatiotemporal relationships of body movements and poses. Once an exercise is performed, the information of pose and movement per frame is stored in the GWR. For every frame, the current pose and motion pair is compared against a predicted output of the GWR, allowing for feedback not only on the pose but also on the velocity of the motion. In a practical scenario, a physical exercise is performed by an expert like a physiotherapist and then used as a reference for a humanoid robot like Pepper to give feedback on a patient’s execution of the same exercise. Since the humanoid robot is mobile, it can be employed in different environments. This approach, however, comes with two challenges. First, the distance from the humanoid robot and the position of the user in the camera’s view of the humanoid robot have to be considered by the GWR as well, requiring a robustness against the user's positioning in the field of view of the humanoid robot. Second, since both the pose and motion are dependent on the body measurements of the original performer, the expert’s exercise cannot be easily used as a reference. This paper tackles the first challenge by designing an architecture that allows for tolerances in translation and rotations regarding the center of the field of view. For the second challenge, we allow the GWR to grow online on incremental data. In order to support multiple persons, catastrophic forgetting has to be closely monitored and evaluated. For evaluation, we developed a new synthetic dataset methodology and created a novel exercise dataset with virtual avatars called the Virtual-Squat dataset. Overall, we claim that our novel architecture based on the GWR can use a learned exercise reference for different body variations through continual online learning, while preventing catastrophic forgetting, enabling for an engaging long-term human-robot interaction with a humanoid robot.
\end{abstract}
\section{Introduction}\label{introduction}
A lack of physical exercise is directly linked to many health issues including obesity, cardiovascular diseases as well as depression and anxiety \citep{booth2011}. Therefore, physical activities are required for a healthy lifestyle \citep{fen2009}. However, performing physical exercises without proper techniques can lead to injuries \citep{gray2015}. As a consequence, supervision by a personal trainer or physiotherapist is of utter importance for people who are unfamiliar with performing physical exercises. Fitness professionals do not only prevent injuries by preventing incorrect techniques, but also increase the effects of the exercise by pushing clients closer to their limits and thereby increasing the overall exercise intensity \citep{de2017}. Despite the benefits, for some people booking a personal trainer is no possibility or a physiotherapist is not available. Therefore, the question arises whether a humanoid robot can act in a supportive manner for fitness professionals to encourage clients to exercise. In order to provide a basic service, the humanoid robot is required to correct improper technique and engage with its user. Thus, the humanoid robot must be able to detect pose and movement of a user and compare it to a learned exercise recalled from memory in order to provide feedback. This requirement comes with multiple challenges. First of all, the humanoid robot should be able to learn an exercise and its corresponding pose and movement pattern. Therefore, the memory cannot be fixed beforehand and has to be expandable. Secondly, the learned sequence of pose and movements of an exercise provided by the initial training might mismatch with the current user's body shape. As a consequence, the memory has to be updated continuously for every new user while maintaining  all previous information. So, the proposed architecture has to counteract catastrophic forgetting. Finally, the humanoid robot has to provide feedback that is valuable, engaging and easy to understand for the user.
\begin{figure}[H]
	\centering
	\includegraphics[width=0.5\textwidth]{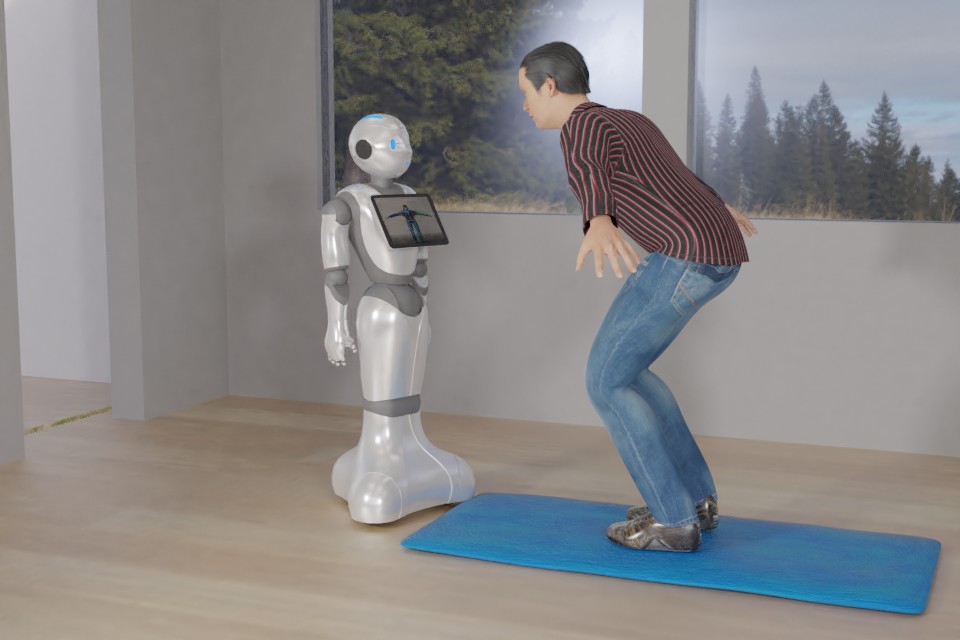}
	\caption{Example scenario, where a user performs physical exercise in front of Pepper, getting feedback via its tablet.}
	\label{fig:pepper-user}
\end{figure}
To tackle these challenges, for one, OpenPose \citep{cao2018}  is utilized as the pose estimation framework. Secondly, for memory and learning, a Grow-When-Required Network (GWR) \citep{marsland2002} with recurrent connection is used. Finally, due to its humanoid form and its tablet as an easy tool for visual feedback, Pepper is selected as the robot. The resulting novelty of this work is twofold. For one, the recurrent variant of the GWR, called Gamma-GWR \citep{parisi2017}, is extended in order to counteract catastrophic forgetting and to store many different variations of body shapes for a pose. We call this network Subnode-GWR. For evaluation, we created a novel exercise dataset based on virtual avatars with differing body shapes on which we are able to achieve an average accuracy of $88\%$ with robustness against rotation and translation. Finally, we employ the architecture together with a humanoid robot in order to lay the foundations for an interactive physical exercise experience.

The rest of the paper is organized as follows. In section \ref{related-work}, an overview of pose trainers and continual learning is given. In section \ref{methodology}, the Subnode-GWR architecture as well as the human-robot scenario are described in detail and evaluated in section \ref{experimental-results}, which will be discussed and concluded in section \ref{conclusion}, where also possible future work is mentioned.
\section{Related Work}\label{related-work}
According to \cite{davis2018} and \cite{mageau2003} performance improvement and stress reduction are coupled with a positive relationship between an athlete and his coach. Consequently, a negative encounter with a coach decreases motivation \citep{bartholomew2009, newsom2005}. This also holds for pose trainers that can be categorized as Smart Coaches. \cite{gamez2020} define Smart Coaching "as a set of smart devices to work independently with the objective of helping people to improve in a specific field".
\subsection{Pose Trainer}\label{pose-trainer}
One can cluster pose trainers into two categories: camera-based or sensor-based. Then, again, both approaches can be fanned out into RGB-D, which is a color image with depth information, or solely RGB for camera-based methods and into classical sensors, e. g., motion sensors and medical systems like electroencephalograms (EEG) or electromyography (EMG) for sensor-based methods. Together with support-vector machines (SVM) proposed by \cite{cortes1995} as a classifier, EEG has been used by \cite{zhang2014} for a rehabilitation training system which has been improved by \cite{ukita2015}. Using an EMG and a SVM, \cite{lee2018} classifies between healthy and sick persons for upper body rehabilitation. In order to detect and analyze protective behavior of patients with chronic pain, \cite{wang2019} use a long short-term memory (LSTM), a recurrent neural network originally  proposed by \cite{hochreiter1997}, that was fed with surface electromyography (sEMG) data in a stand-to-sit-to-stand scenario.

For camera-based methods, many approaches in the health domain make use of the infrared camera Microsoft Kinect. One of its main advantages is its integrated pose estimation. \cite{ukita2014} classify binary the pose of 3D skeletons acquired from a Kinect with an SVM as correct or wrong. \cite{pullen2018} and \cite{trejo2018} classify postures in yoga obtained by Kinects. For weight-lifting, \cite{parisi2015} predict motion patterns with a self-organizing network and compare them with the real-time poses estimated by a Kinect. While the Kinect is easy to use and has a built-in pose estimation based on depth information, its estimation is not very accurate in comparison to current deep-learning  human pose estimation approaches.

According to \cite{zheng2020}, in general, human pose estimation is split up into 2D and 3D pose estimation. In 2D human pose estimation, key points that correspond to the two-dimensional spatial location of each joint in an image are extracted, whereas in 3D estimation also depth information is retrieved. In a next step, one can distinguish between single-person and multi-person detection in the 2D domain. The two mainly used deep learning methods for single person detection are regression and body part detection \citep{zheng2020}. In regression approaches, the pipeline takes an image as an input and outputs key points in an end-to-end manner. Therefore, a direct mapping from the input image to the 2D pose is learned. For body part detection, the pipeline consists of two steps. First, for each body part a heatmap that indicates the probability for a key point to match the individual joint location is created. In a second step, the key points and the corresponding body parts are put in relation to each other and the overall pose is generated. The shift from traditional approaches in human pose estimation towards deep learning was pushed by \cite{toshev2014} and their multi-stage convolutional neural network (CNN) regressor DeepPose. Since then, human pose estimation frameworks have improved steadily and most of today's best-performing architectures are based on the body part detection approach. In contrast to single person estimation, multi-person pose estimation faces the challenge of having multiple key points for one joint type that have to be matched to the correct person. Therefore, the idea quickly raised to use a person or object detector like YOLO by \cite{redmon2016} first in order to receive cropped images where just one person is visible and apply one of the single-person methods. However, this comes with a major drawback, since the accuracy of the human pose estimation depends heavily on the performance of the involved person or object detector. As a solution, bottom-up methods have been developed. One of them is called OpenPose by \cite{cao2018}. As an architecture, it consists of two multi-stage CNNs. For preprocessing, a VGG convolutional network, originally developed by \cite{simonyan2014} extracts the features of the inputted image. From these feature maps, the first CNN in an OpenPose architecture computes so-called part affinity field maps (PAFs), that indicate the connection between the joints to form the body part. These PAFs together with the original image features from VGG19 are used in a second CNN to compute the joint locations for each body part. Finally, these heatmaps are used to match the body parts to the correct person in the scene by applying bipartite matching. The pose estimation is superposed on the original image as a skeleton figure.

In general, convolutional neural networks are a powerful tool for pose estimation, as e. g. \cite{kamel2019} show, who designed a convolutional neural network to provide real-time feedback on Tai Chi poses. \cite{liao2020} propose a framework that gives a metric for quantifying movement performance. They also introduce scoring functions which map the metric into numerical scores of movement quality. To achieve this, a deep neural network is developed, which generates quality scores for inputted movements. The neural network receives the joint coordinates as its input that is split into multiple individual body parts and their joint coordinates. The input data for each body part is arranged into temporal pyramids, where multiple scaled versions of the movement repetitions are processed with 1D convolutions and concatenated. Then, the concatenated output is fed into a series of LSTM layers in order to model temporal correlations in learned representations. Finally, a linear layer outputs a movement quality score. Another smart coach proposed by \cite{zou2018} uses the regional multi-person pose estimation (RMPE) framework developed by \cite{fang2017} to extract poses from video to generate feedback on the physical exercise performance of users. Recently, \cite{ota2020} verify OpenPose reliability and accuracy on motion analysis for bilateral squats. Therefore, we select OpenPose as our framework to use, since we also analyze a variation of squats as described in section \ref{experimental-results}. Furthermore, it allows the usage of the humanoid robot Pepper with its built-in RGB camera without requiring an additional depth camera, which increases the usability of our approach.  However, as aforementioned in section \ref{introduction}, the problem still arises how to adapt to different body sizes and variations that significantly mismatch with the trained-on key points. As a solution, we develop continual learning schemes for our architecture, that allow for adaption to unknown body shapes.

\subsection{Continual Learning}\label{continual learning}
Continual learning, also referred to as lifelong learning, is deeply integrated into the learning of humans, such that they develop their cognitive and motoric skills based on novel experiences and repetition of already acquired knowledge over their lifespan (see \citep{parisi2019} for a review). Herein also lies the main challenge of continual learning: catastrophic forgetting. Catastrophic forgetting describes the process where previously learned tasks or information are overwritten by novel knowledge \citep{mcclelland1995}. This issue finds itself also in the human brain, where it is expressed as a stability-plasticity dilemma \citep{mermillod2013}. The neural structure in brain areas have to be able to change in order to integrate new information while keeping already acquired knowledge intact. This neurosynaptic plasticity is essential for human learning and is at its highest during early development, where the input is dominated by novel sensorimotor experiences \citep{parisi2019}. While the brain stays plastic over a lifetime, it becomes less prominent over time when stable neural connections have been established \citep{hensch1998}. The underlying mechanism for controlling the plasticity and stability are based on the presynaptic and postsynaptic strength, which was discovered by \cite{hebb1949}. As soon as one neuron is excited by an external stimulus, it activates neurons connected to it. The degree of activation depends on the connection's strength that is updated based on the presynaptic and post-synaptic activity. While Hebbian plasticity is the basis for neurosynaptic adaptation, the complementary learning system (CLS) theory articulated by \cite{mcclelland1995} is the scheme that drives the learning and memorization process. The hippocampus acts as an episodic memory that is highly plastic and therefore learns fast. On the other hand, the neocortex learns slowly and as a consequence acts as a long-term storage for information. In order to store knowledge and counteract catastrophic forgetting, the structure of the neocortex only changes after receiving similar input over a longer time span. Therefore, the hippocampus replays episodic events to the neocortex, which will incorporate the knowledge given the repeating activation of similar structures.

It comes with no surprise that these brain mechanisms have been implemented in lifelong machine learning approaches. One basic approach stems from \cite{kohonen1990} and is called a self-organizing map (SOM). It has fixed structures consisting of nodes that represent neurons in the brain. To each node, a weight is assigned that defines its place in the input space and therefore also in the lattice of the self-organizing map. This lattice is trained by finding the best-matching node with the least distance to an input sample. The weight is updated according to the difference between the input sample and the node's distance. Also, neighboring node weights that are connected to the best-matching node are updated. As a consequence, the lattice of the self-organizing map deforms until the average distance to all input samples is minimized. However, since self-organizing maps are fixed in their number of nodes and thus in their dimensions, they are not suitable for multitask challenges in the lifelong learning context. Therefore, self-organizing maps have been extended by, e. g., Growing Neural Gases \citep{fritzke1995}. They allow for nodes to be deleted and added. The addition of nodes though occurs after a fixed amount of iterations, which forbids a dynamic growth based on the need for new nodes to represent the input space.

Grow-When-Required (GWR) networks by \cite{marsland2002} overcome this issue by allowing nodes to be added dynamically whenever the best-matching node's activity is lower than a predefined threshold. While Grow-When-Required networks are able to learn static input, they lack the possibility to store temporal information between the input samples. Therefore, recurrences are introduced in the Gamma-GWR from \cite{parisi2017} as context vectors that are additionally stored for every node. They are based on the ideas of the Merge SOM  architecture proposed by \cite{strickert2005}, where context descriptors capture the activity of the self-organizing map for a given time step. As a consequence, the distant function of the Gamma-GWR not only depends on the weights of a node but also on its context that is based on the activation of previous time steps. Thus, time sequences can be incorporated in the structure of the Gamma-GWR allowing to learn, e. g., spatiotemporal sequences.  To this comes  the caveat that the input sample in every time step has to be unique, since otherwise nodes link to themselves, which results in a loop for the time sequence. This is not an issue with the  Episodic-GWR \citep{parisi2018} that directly stores the predecessor of a node and does not allow for a node to be its own predecessor. This leads to a loss of information for a time sequence, e. g., a physical exercise where a pose has to be held for a longer period of time, which is why we extend this approach with our Subnode-GWR.
\section{Physical Exercise Scenario with a Humanoid Robot}\label{methodology}
In our design, the humanoid robot Pepper from Softbank Robotics acts as a motivator and trainer for the user performing a physical exercise, which is demonstrated in figure \ref{fig:pepper-user}. Pepper has been designed for human-robot interaction especially, featuring built-in speech and face recognition through their NAOqi-API. In its head, microphones, speakers and cameras are installed and it can move on wheels that are integrated in its triangular base with multiple environment sensors for navigation. Pepper is equipped with tactile sensors in its hands and head. Overall, 17 joints can be manipulated for expressive gestures and visual feedback can be given on its tablet that is mounted on its chest. While Pepper in its core is designed as a humanoid robot, it has no explicit gender, which is also expressed in Pepper's androgynous voice. When using it for different clients, this is advantageous as studies show that persons are biased towards robots expressing a gender \citep{siegel2009, tay2014}.
\subsection{Human-Robot Interaction}
We use Pepper's tablet to mirror the real-time video feed from Pepper's camera in its head. To do so, the video feed from the camera is streamed to a computer, where it is processed by OpenPose in real-time. The extracted key points are drawn as a skeleton figure on the frame. We also embed the visual feedback into this skeleton figure. Therefore, we compare the difference in the joint's key points between the estimation from OpenPose and the stored information in our Subnode-GWR per video frame. If the error is larger than a predefined threshold, we render the corresponding joint in the skeleton as red, indicating that the current joint's position is wrong. Otherwise, the joint is drawn in green reflecting the correct positioning of the joint. We stream the frame with the user and the superposed skeleton figure to a local web server, that can be accessed by Pepper, which is then displayed on its tablet, giving real-time, intuitive and supportive feedback to the user in front of the Pepper robot. The scenario is shown in figure \ref{fig:pepper-user}.

\begin{figure}[h]
	\centering
	\includegraphics[width=0.75\textwidth]{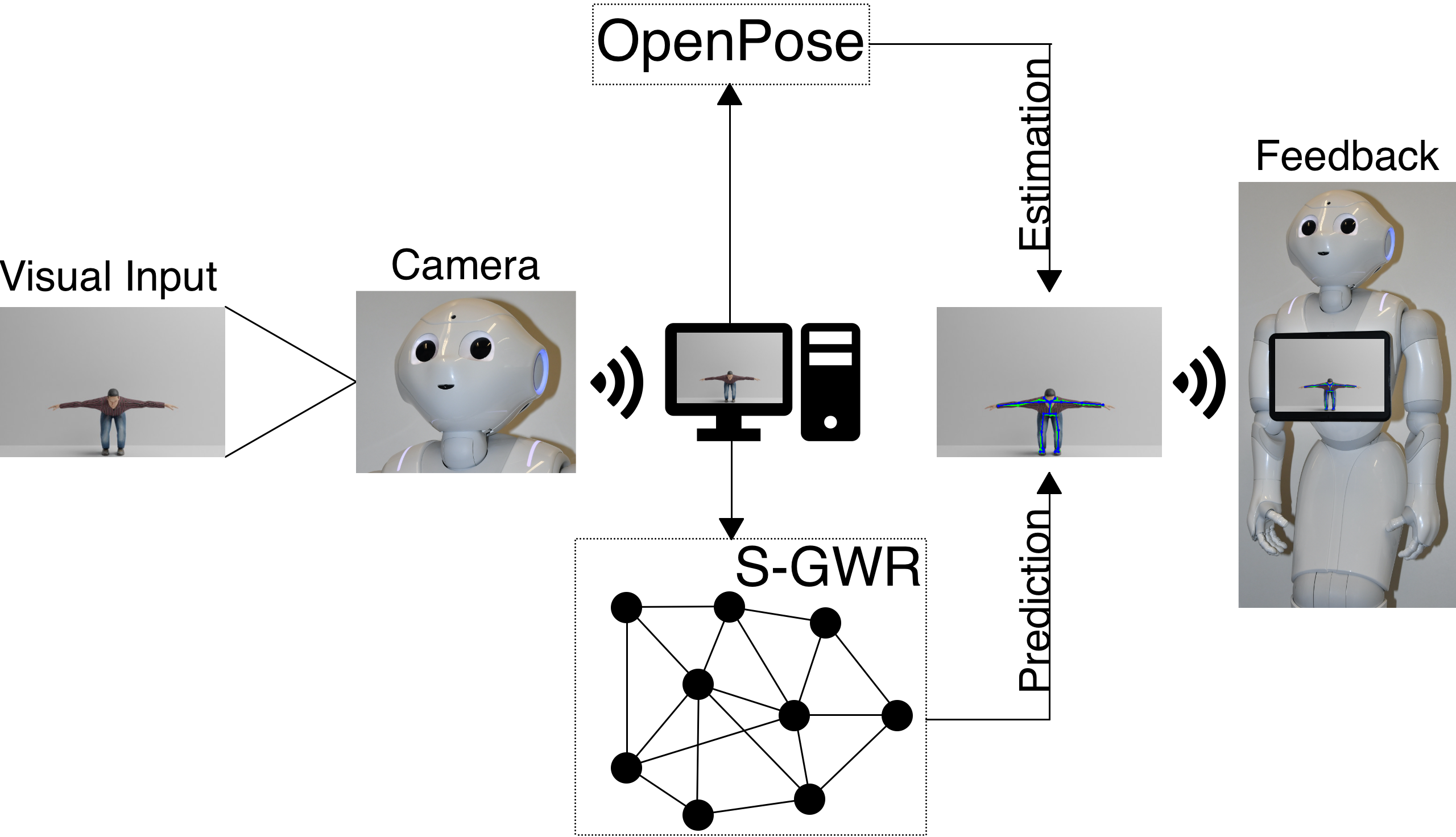}
	\caption{Flowchart that demonstrates the overall scenario with the Pepper and architectures involved.}
	\label{fig:flowchart}
\end{figure}

Additionally, Pepper should react accordingly with verbal and gestural feedback, e. g., praising the user if he/she has performed well, hinting at possible areas of improvement if there is a dominant issue and motivating the user to continue exercising. For gestural feedback, Pepper's movement should be restricted to its arms and hands. This is due to the fact, that we record the user in front of Pepper through the integrated camera in its head and need to minimize Pepper's head movement. In order to correctly process the poses trained on and embedded in the Subnode-GWR, the user is asked by the Pepper to position him-/herself in the camera's field of view such that no key points are cut off. Hence, the real-time estimation by OpenPose is shown on Pepper's tablet, making it clearly visible to the user whether he/she is positioned accurately. On top of that, we expect that our Subnode-GWR works within a tolerance of 5 degrees in rotation and 5 centimeters in translation, which we evaluate in section \ref{experimental-results}. The overall data processing is illustrated in figure \ref{fig:flowchart}.
\subsection{Subnode-GWR}\label{subnode-gwr}
In order to train the Subnode-GWR, a video of a physical exercise that has been performed correctly is processed by OpenPose in order to receive poses as key points for each frame. These key points are normalized according to the image dimensions of a frame and fed into the architecture as training samples. The Subnode-GWR is initialized at first with two nodes that are randomly selected from the number of samples. For an input sample, the distance to each node in the Subnode-GWR is calculated as
\begin{equation}
	d_{j}=\alpha_{0}\left\Vert\mathbf{x}(t)-\mathbf{w}_j\right\Vert_{2}+\sum_{k=1}^{K}\alpha_k\left\Vert\mathbf{C}_k(t)-\mathbf{c}_{j,k}\right\Vert_{2}.
	\label{eqn:distance}
\end{equation}
In equation \ref{eqn:distance}, $\mathbf{x}(t)$ refers to the sample at time step $t$ and $\mathbf{w}_j$ to the weight vector of node $j$. $\mathbf{c}_{j,k}$ is the context of the $j$th node. It incorporates information of the previous activation in the map up to $k$ time steps. $\mathbf{C}_k(t)$ is the context descriptor that is computed as
\begin{equation}
	\mathbf{C}_k(t)=\beta\cdot\mathbf{w}_b^{t-1}+(1-\beta)\cdot\mathbf{c}_{b,k-1}^{t-1},
	\label{eqn:context_descriptor}
\end{equation}
where $b$ denotes the best-matching unit (BMU) with the smallest distance of all nodes according to
\begin{equation}
	b=\arg \min_{j\in V}(d_j).
	\label{eqn:bmu}
\end{equation}
The factors $\alpha_{0}$ and $\alpha_k$ are used to balance the influence between the weight vector and the context on the distance to an input sample. In the next step, the activity of the network $a(t)$ is computed based on the BMU as follows:
\begin{equation}
	a(t)=\exp(-d_b),
	\label{eqn:activity}
\end{equation}
which, as a consequence, allows a maximal activity of $1$. If the activation $a(t)$ is lower than a predefined threshold $a_t$, one criteria is met to add a new node to the network. The other criteria is the node's habituation counter $h_j\in[0, 1]$, which allows the nodes to be trained properly, before expanding the network. Being initialized with $h_j=1$, each node's habituation counter is decreased towards $0$ over time whenever a BMU has fired. The habituation counter $h_b$ for the BMU and $h_n$ for the neighboring nodes is reduced by
\begin{equation}
	\Delta h_i=\tau_i\cdot\kappa\cdot(1-h_i)-\tau_i,
	\label{eqn:habituation}
\end{equation}
where $i\in\{n,b\}$ and with $\tau_i$ and $\kappa$ regulating the speed of  decrease. According to \cite{parisi2018}, $h_b$ should usually decrease faster than $h_n$, thus $\tau_b$ and $\tau_n$ are selected such that $\tau_b>\tau_n$. For the case that $h_b$ as well as $a(t)$ are less than $h_t$ and $a_t$ respectively, a new node $r$ is added to the network by removing the connection between the best-matching and second-best-matching node and connect both to the added node. Its connection age is set to $0$. Its weight and context vector are computed as
\begin{equation}
\label{eqn:new_node}
\begin{aligned}
	\mathbf{w}_r&=0.5\cdot(\mathbf{x}(t)+\mathbf{w}_b),\\
	\mathbf{c}_{r, k}&=0.5\cdot(\mathbf{C}_k(t)+\mathbf{c}_{b,k}).
\end{aligned}
\end{equation}
For the case that the activity of the network $a(t)$ and/or the habituation counter $h_b$ are greater than or equal to the thresholds $a_t$ and/or $h_t$, the BMU $b$ and its neighboring nodes are updated as follows
\begin{equation}
	\label{eqn:update_node}
	\begin{aligned}
		\Delta\mathbf{w}_i&=\epsilon_i\cdot h_i\cdot(\mathbf{x}(t)-\mathbf{w}_i),\\
		\Delta\mathbf{c}_{i, k}&=\epsilon_i\cdot h_i\cdot(\mathbf{C}_k(t)-\mathbf{c}_{i,k}),
	\end{aligned}
\end{equation}
with $i\in\{n, b\}$ and where $\epsilon_i$ are constant learning rates that are usually selected as $\epsilon_b>\epsilon_n$. Also, all connections that end in the BMU $b$ are aged by one and will be removed if their age is larger than a predefined threshold $\mu_{max}$. Finally, all nodes that are not connected to any other node are considered dead and are removed. In contrast to the Episodic-GWR, the information about the successor of a node is not encoded in a matrix $P$, where each connection between nodes is stored and increased by $1$ if two nodes are activated consecutively. While this allows to recall a trajectory of activation by selecting each node's most frequent consecutively activated BMU, it forbids to select  itself as its own successor according to
\begin{equation}
	\label{eqn:episodic_p}
	v=\arg\max_{j\in V\backslash i}P_{(i,j)}.
\end{equation}
We modify in our  Subnode-GWR architecture $P_{(i,j)}$ to become $P_{e_i}$, where each row $e_i$ in $P_{e_i}$ resembles one physical exercise that the network is supposed to recall. The row itself consists of the best matching units $b_{i, t}$  in consecutive order as they were activated during the last epoch of training on a physical exercise:
\begin{equation}
	\label{eqn:store_exercise}
	P_{e_i}=\begin{bmatrix}e_0 \\e_1 \\e_2 \\\vdots \\e_i\end{bmatrix},
	e_i=\begin{bmatrix}b_{i,0}, b_{i,1}, b_{i,2}, ..., b_{i,t}\end{bmatrix}.
\end{equation}
There are two advantages to this approach. On the one hand, since Gamma-GWRs solely rely on context to determine a node's successor, they tend to loop in their prediction if a node references to itself. On the other hand, Episodic-GWRs, according to equation \ref{eqn:episodic_p}, forbid nodes to be their own successor at all. This limits the capability of the network to learn physical exercises that require to hold a pose for some frames. These issues are resolved by the modifications described in equation \ref{eqn:store_exercise}, which allow for nodes to precede themselves without looping and thus making it possible to store physical exercises, where one pose spans over a longer time frame. The complete algorithm is also depicted in algorithm \ref{alg:sgwr-algorithm}.
\begin{algorithm}
	\caption{Training of  Subnode-GWR (S-GWR)}
	\label{alg:sgwr-algorithm}
	\begin{algorithmic}[1] % The number tells where the line numbering should start
		\STATE Create first two nodes $V=\{\mathbf{w_1}, \mathbf{w_2}\}$ initialized with first two samples $\mathbf{x}(0), \mathbf{x}(1)$  from input $\mathbf{X}$ and empty context vectors $\mathbf{c}_i^k$ for $k=1, ..., K, i=1,2$.
		\STATE Initialize connection set as $E=\emptyset$.
		\STATE Initialize global context $\mathbf{C}_k(t)$ for $k=1, ...,K$ as empty.
		\STATE Initialize exercise matrix $P_{e_i}=P_{e_i}\cup e_0$ with $e_0=\emptyset.$
		\FOR{$n_{epoch} < n_{epoch,max}$}
		\FOR{$n_{iter} < \left\vert X\right\vert$}
		\STATE Generate input sample $\mathbf{x}(t)$.
		\STATE Compute distance for every node:\newline
		$d_{j}=\alpha_{0}\left\Vert\mathbf{x}(t)-\mathbf{w}_j\right\Vert_{2}+\sum_{k=1}^{K}\alpha_k\left\Vert\mathbf{C}_k(t)-\mathbf{c}_{j,k}\right\Vert_{2}$
		\STATE Select best and second-best matching neurons:\newline$b=\arg \min_{j\in V}(d_j), s=\arg \min_{j\in V/\{b\}}(d_j)$
		\STATE Update global context:\newline
		$\mathbf{C}_k(t)=\beta\cdot\mathbf{c}_{b(t-1), k}+(1-\beta)\cdot\mathbf{c}_{b(t-1), k-1}$ for $k=1, ..., K$.
		\IF{$E \neq E\cup\{(b, s)\}$}
		\STATE Set $E=E\cup\{(b, s)\}$.
		\STATE Set age of connection to $0$.
		\ENDIF
		\STATE Compute activity for BMU:\newline
		$a(t)=\exp(-d_b)$.
		\IF{$a(t)<a_t$ \AND $h_b<h_t$}
		\STATE Add a new neuron $r$  such that $V=V\cup\{r\}$ with:\newline
		$\begin{aligned}
			\mathbf{w}_r&=0.5\cdot(\mathbf{x}(t)+\mathbf{w}_b),\\[-1\jot]
			\mathbf{c}_{r,k}&=0.5\cdot(\mathbf{C}_k(t)+\mathbf{c}_{b,k}).
		\end{aligned}$
		\STATE Set habituation counter $h_r$ to $1$.
		\STATE Change connections between neurons:\newline
		$E=E\cup\{(r,b),(r,s)\}$ and $E=E/\{(b,s)\}.$
		\ELSE
		\STATE Update weight  and context of BMU $b$ and neighbors $n$ $(i=\{b, n\})$:\newline
		$\begin{aligned}
			\Delta\mathbf{w}_i&=\epsilon_i\cdot h_i \cdot(\mathbf{x}(t)+\mathbf{w}_i),\\[-1\jot]
			\Delta\mathbf{c}_{i,k}&=\epsilon_i\cdot h_i\cdot(\mathbf{C}_k(t)+\mathbf{c}_{i,k}).
		\end{aligned}$
		\ENDIF
		\STATE Increase age of all connections that end in BMU $b$ by $1$.
		\STATE Reduce habituation counter $h_i$ for BMU $b$ and neighbors $n$ $(i=\{b, n\})$:\newline
		$\Delta h_i=\tau_i\cdot\kappa\cdot(1-h_i)-\tau_i.$
		\STATE Purge edges with ages larger than $\mu_{max}$ and remove nodes without connections.
		\IF{$n_{epoch}=n_{epoch,max}-1$}
		\STATE Append BMU $b$ to $e_0$ such that $e_{0,i}=b $ for $(i={0, ..., \left\vert X\right\vert}-1)$
		\ENDIF
		\STATE Increase $n_{iter}$ by 1.
		\ENDFOR
		\STATE Increase $n_{epoch}$ by 1.
		\ENDFOR
	\end{algorithmic}
\end{algorithm}
After training, the Subnode-GWR can recall the pattern of poses and motion vectors for the trained physical exercise. However, the network is tuned for the body dimensions it has been trained on, which limits its ability to be used for analyzing movements from other users. Thus, the primary extension from the Subnode-GWR to the Gamma-GWR stems from the necessity to apply the trajectory of BMUs, that are stored in $P_{e_i}$ and resemble a physical exercise, to different body shapes and variations following from mismatching, e. g., age, gender and/or general appearance of the performer of a sample exercise. Therefore, the architecture needs to learn continuously adapting to a first-time user. Hence, we integrate subnodes to existing nodes. Their weight vector and context vector are computed as:
\begin{equation}
	\label{eqn:new_subnode}
	\begin{aligned}
		\mathbf{w}_{i,j,l} &=\mathbf{w}_{i,j} + (\mathbf{x}(t) + \mathbf{w}_{i,j}),\\
		\mathbf{c}_{i,j,l,k}&=\mathbf{c}_{i,j,k}.
	\end{aligned}
\end{equation}
For a given physical exercise, that has been previously trained and henceforth, a trajectory $e_i$ exists, we extend each BMU $b_{i,j}$ that currently mismatches the input $\mathbf{x}(t)$ with a subnode. To do so, according to equation \ref{eqn:new_subnode}, the difference between the input $\mathbf{x}(t)$ and the weight vector of the current BMU is added to the BMU. Since the entry point to the subnodes is always the parent node, the context $c_{i,j,k}$ is simply copied. This allows the Subnode-GWR to easily adapt to new unseen body shapes and variations, while keeping the trajectory of BMUs that maps the physical exercise intact and prevents loss of knowledge about previous body shapes. We use $P_{e_i}$ to compare the real-time pose estimation of the user from OpenPose with the weight vector of the current BMU $b_{i,j}$ directly or one of its subnodes $b_{i,j,l}$ for exercise $e_i$ if the error on the first frame is lower. The distance between the actual and supposed pose is computed as
\begin{equation}
	\label{distance}
	d_{pose}=\left\Vert\mathbf{x}(t)-w_{i,j,l}\right\Vert_2.
\end{equation}
We use $d_{pose}$ to display the joint-wise error in the current pose compared to the supposed pose of a physical exercise in our human-robot interaction, allowing for precise feedback to the user. Should $d_{pose}$, however, be larger than a predefined threshold $d_{t,learning}$ on the first frame, the continual learning scheme is triggered, where for each BMU $b_{i,j}$ in trajectory $e_i$ a subnode is created corresponding to the current input pose $\mathbf{x}(t)$. Also, the user is asked to perform the physical exercise once as a baseline. It is important to note that for this step, a fitness professional is advised, since all feedback following is, due to the architecture of the Subnode-GWR, established on this initial performance. Else, if $d_{pose}<d_{t,learning}$ the training with the Pepper is executed as described beforehand. The algorithm supporting continual learning is shown in algorithm \ref{alg:sgwr-continual-algorithm}.
\begin{algorithm}
	\caption{Continual Learning of  Subnode-GWR (S-GWR)}
	\label{alg:sgwr-continual-algorithm}
	\begin{algorithmic}[1]
		\STATE Compare $\mathbf{x}(t)$ from real-time input with first node in $e_i$ of $P_{e_i}$\newline
		$d_{pose}=\left\Vert\mathbf{x}(t)-w_{i,j,l}\right\Vert$.
		\STATE Select node $b_{i,j}$ with smallest distance to $\mathbf{x}(t)$.
		\IF{$d_{pose} > d_{t,learning}$}
			\FOR{$j<\left\vert e_{i}\right\vert$}
				\STATE Create subnode $l$ for node $b_{i,j}$ in $e_i$ with:\newline
				$\begin{aligned}
					\mathbf{w}_{i,j,l}&=\mathbf{w}_{i,j}+(\mathbf{x}(t)-\mathbf{w}_{i,j}),\\[-1\jot]
					\mathbf{c}_{i,j,l,k}&=\mathbf{c}_{i,j,k}
				\end{aligned}$
			\ENDFOR
		\ELSE
			\STATE Use trajectory of BMUs $b_{i,j}$ without creating subnodes for phyiscal exercise $e_i$.
		\ENDIF
	\end{algorithmic}
\end{algorithm}
\section{Experiments on Subnode-GWR Performance and Robustness}\label{experimental-results}
\begin{figure}[!htbp]
	\setcounter{subfigure}{0}
	\centering
	\begin{minipage}[b]{\textwidth}
		\includegraphics[width=0.195\textwidth]{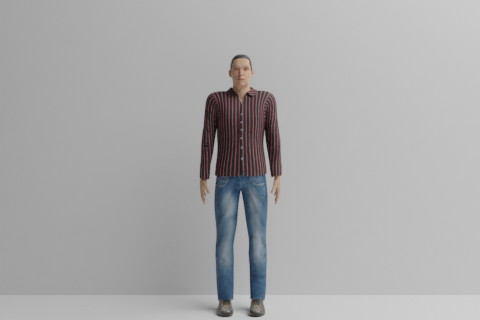}
		\includegraphics[width=0.195\textwidth]{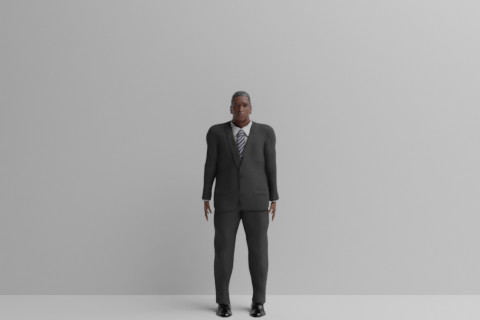}
		\includegraphics[width=0.195\textwidth]{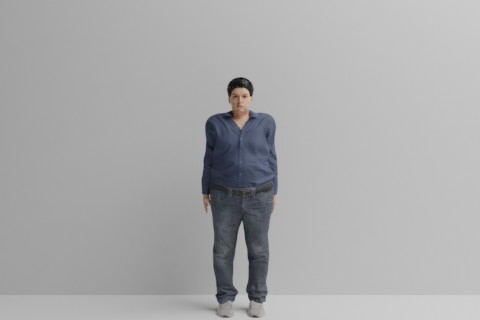}
		\includegraphics[width=0.195\textwidth]{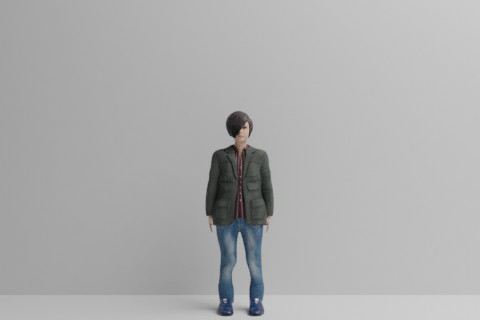}
		\includegraphics[width=0.195\textwidth]{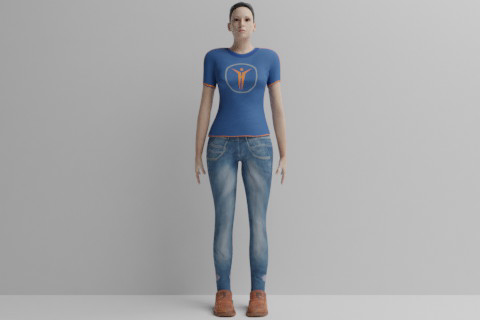}
	\end{minipage}
	\vskip\baselineskip
	\begin{minipage}[b]{\textwidth}
		\includegraphics[width=0.195\textwidth]{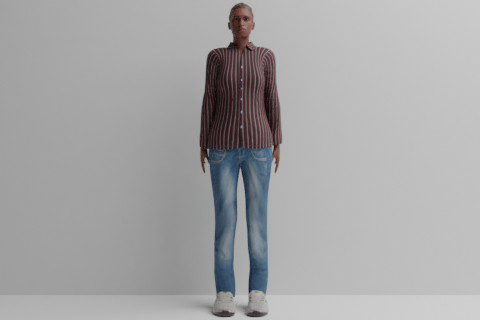}
		\includegraphics[width=0.195\textwidth]{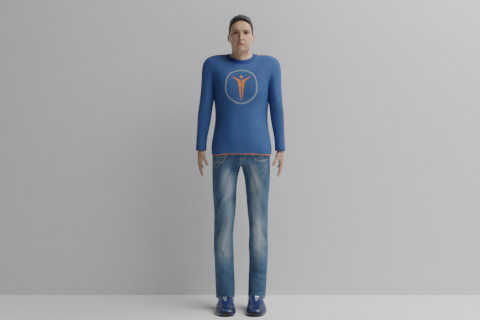}
		\includegraphics[width=0.195\textwidth]{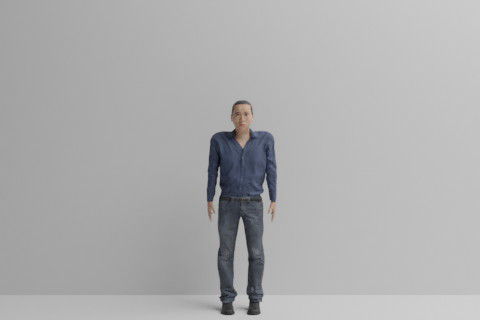}
		\includegraphics[width=0.195\textwidth]{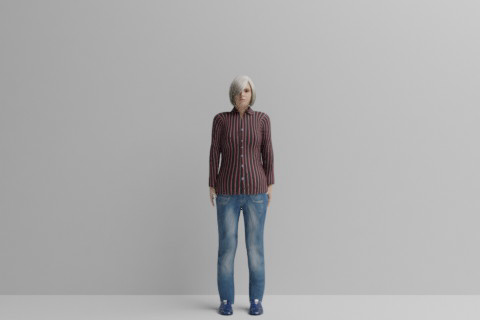}
		\includegraphics[width=0.195\textwidth]{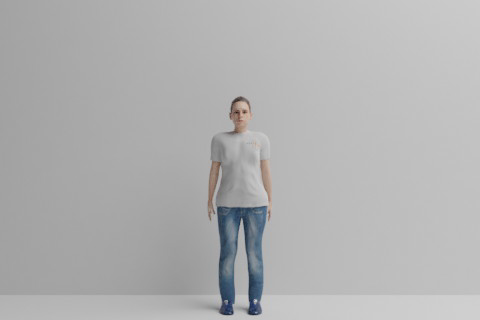}
	\end{minipage}
	\caption{Overview of the ten virtual avatars in the Virtual-Squat dataset, differing in height, weight, clothes, color of skin and gender.}
	\label{fig:avatar-comparison}
\end{figure}

In order to evaluate our approach, the Virtual-Squat dataset\footnote{\url{https://www.inf.uni-hamburg.de/en/inst/ab/wtm/research/corpora.html}}  was created using the Blender open-source 3D creation suite\footnote{\url{https://www.blender.org/}} and the MakeHuman open-source creation tool for virtual humans\footnote{\url{http://www.makehumancommunity.org/}}. The dataset consists of ten different avatars (shown in figure \ref{fig:avatar-comparison}). The avatars have randomized heights, weights, body shapes, clothing, skin colors, hairstyles and hair colors to evaluate the robustness of the pose estimation to superficial visual properties of the avatars and the robustness of the exercise analysis to different proportions.

For the physical exercise, we selected a squat where, instead of keeping the arms straight in front of the body as is typical for this exercise, we opted for side arm raises. The selected exercise offers multiple advantages. First of all, for the user him-/herself, this movement strains muscles that are required for essential daily activities, e. g., lifting and sitting as well as sports movements \citep{myer2014}. Secondly, it is challenging for human pose estimation, since all joints are moved. Thirdly, it is a cycling exercise with two halting points (standing upright, sitting down), which is demanding on the pose prediction (see figure \ref{fig:first-experiment}). Finally, the camera view is monocular and the pose estimation two-dimensional. Therefore, the camera cannot capture the physical exercise extending into the depth while squatting down, increasing the difficulty.
\begin{figure}[!htbp]
	\setcounter{subfigure}{0}
	\centering
	\begin{minipage}[b]{0.195\textwidth}
		\includegraphics[width=\linewidth]{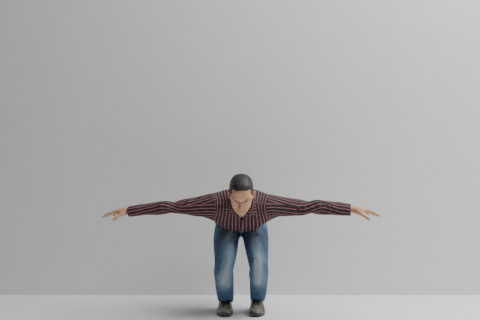}
		\subcaption{Correct}
		\label{fig:avatars-exercises-correct}
	\end{minipage}
	\begin{minipage}[b]{0.195\textwidth}
		\includegraphics[width=\linewidth]{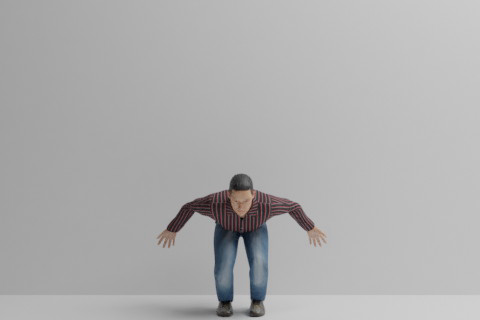}
		\subcaption{Arms not raised}
		\label{fig:avatars-exercises-arms}
	\end{minipage}
	\begin{minipage}[b]{0.195\textwidth}
		\includegraphics[width=\linewidth]{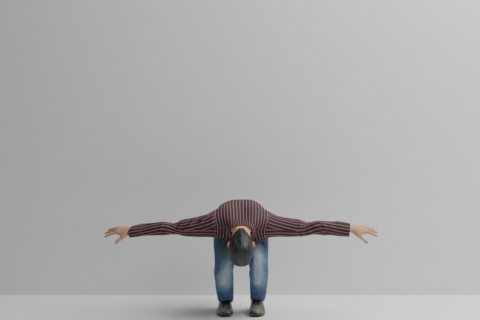}
		\subcaption{Neck strained}
		\label{fig:avatars-exercises-head}
	\end{minipage}
	\vskip\baselineskip
	\begin{minipage}[b]{0.195\textwidth}
		\includegraphics[width=\linewidth]{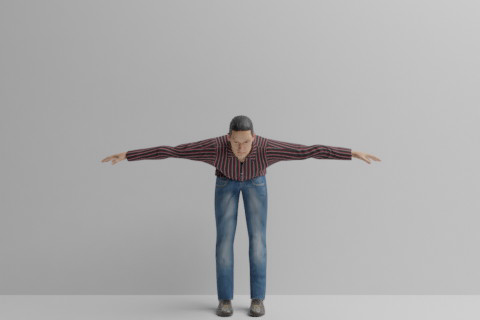}
		\subcaption{Knees not bend}
		\label{fig:avatars-exercises-legs}
	\end{minipage}
	\begin{minipage}[b]{0.195\textwidth}
		\includegraphics[width=\linewidth]{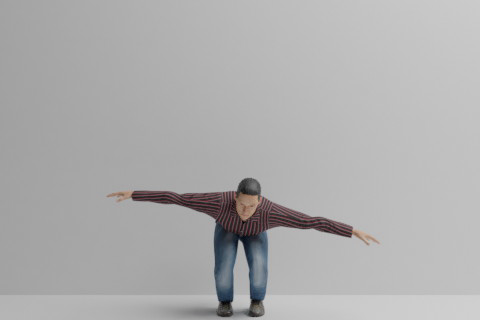}
		\subcaption{Upper body tilted}
		\label{fig:avatars-exercises-side}
	\end{minipage}
	\begin{minipage}[b]{0.195\textwidth}
		\includegraphics[width=\linewidth]{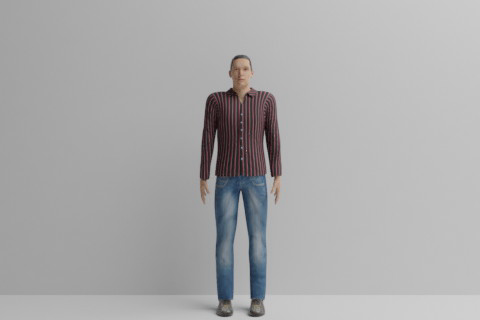}
		\subcaption{Too fast}
		\label{fig:avatars-exercises-speed}
	\end{minipage}
	\caption{All common errors rendered for virtual avatar 01 in comparison to the correct execution. Shown is the 50th frame of the 100-frame videos.}
	\label{fig:avatar-exercises}
\end{figure}

Each avatar is repeating the same exercise (squat) in one correct and five incorrect ways. The squat exercise was animated by hand using a video recording of a correctly performed exercise. The five incorrect executions model frequent errors during the exercise. Correct and incorrect exercises were then applied as animations for the ten avatars, which is illustrated in figure \ref{fig:avatar-exercises}. Note that the virtual avatars offer the benefit of creating incorrect exercise executions, which could be harmful to real persons and cannot be recorded with participants. Each exercise is recorded over 100 frames in 480x320 resolution (Pepper's camera resolution) using the Cycles rendering engine. Furthermore, to simulate imperfect alignment between Pepper and its interaction partner, each execution of the exercise was recorded in four different ways: 1) with the avatar centered in the image and facing straight ahead; 2) with the avatar rotated by 5 degrees clockwise; 3) with the avatar translated by 5 centimeters to the left and 4) with the avatar both rotated and translated. In total, the dataset contains 10 different avatars performing 6 different exercise executions with 4 different rotations and translations, which equals 240 exercise videos.
\subsection{Motion Prediction With Gamma-GWR}
In our first experiment, we evaluate the Gamma-GWR motion prediction capabilities. We therefore process virtual avatar 01 that performs the squat with OpenPose to extract the key points. Then, we train the Gamma-GWR on these key points. For all architectures, we chose the parameters as noted in table \ref{tab:gwr-parameters}.
\begin{table}[!htbp]
	\centering
	\caption{Network parameters used for Gamma-GWR, Episodic-GWR and Subnode-GWR used in and optimized on all experiment results.}
	\label{tab:gwr-parameters}
	\begin{tabular}{|
			>{\columncolor[HTML]{EFEFEF}}c |c|c|c|}
		\hline
		\cellcolor[HTML]{C0C0C0}\textbf{Parameters} & \cellcolor[HTML]{C0C0C0}\textbf{Gamma-GWR} & \cellcolor[HTML]{C0C0C0}\textbf{Episodic-GWR} & \cellcolor[HTML]{C0C0C0}\textbf{Subnode-GWR}       \\ \hline
		\textbf{$\alpha$}                              & \multicolumn{3}{c|}{0.5}                                                                                                                        \\ \cline{1-1}
		\textbf{$\beta$}                               & \multicolumn{3}{c|}{\cellcolor[HTML]{EFEFEF}0.5}                                                                                                \\ \cline{1-1}
		\textbf{$c_k$}                               & 5                                          & \multicolumn{2}{c|}{1}                                                                             \\ \cline{1-1}
		\textbf{$\epsilon_b$}                             & \multicolumn{3}{c|}{\cellcolor[HTML]{EFEFEF}0.2}                                                                                                \\ \cline{1-1}
		\textbf{$\epsilon_i$}                             & \multicolumn{3}{c|}{0.001}                                                                                                                      \\ \cline{1-1}
		\textbf{$\kappa$}                              & \multicolumn{3}{c|}{\cellcolor[HTML]{EFEFEF}1.05}                                                                                               \\ \cline{1-1}
		\textbf{$\tau_b$}                             & \multicolumn{3}{c|}{0.3}                                                                                                                        \\ \cline{1-1}
		\textbf{$\tau_n$}                             & \multicolumn{3}{c|}{\cellcolor[HTML]{EFEFEF}0.1}                                                                                                \\ \cline{1-1}
		\textbf{$a_t$}                               & \multicolumn{3}{c|}{0.99}                                                                                                                       \\ \cline{1-1}
		\textbf{$h_t$}                               & \multicolumn{3}{c|}{\cellcolor[HTML]{EFEFEF}0.3}                                                                                                \\ \cline{1-1}
		\textbf{$\mu_{age}$}                            & \multicolumn{3}{c|}{20}                                                                                                                         \\ \cline{1-1}
		\textbf{$\mu_{size}$}                           & \multicolumn{3}{c|}{\cellcolor[HTML]{EFEFEF}200}                                                                                                \\ \cline{1-1}
		\textbf{$d_{t,pose}$}                            & \multicolumn{3}{c|}{5 pixel (normalized: 0.04)}                                                                                                 \\ \cline{1-1}
		\textbf{$d_{t,learning}$}                           & \multicolumn{2}{c|}{\cellcolor[HTML]{EFEFEF}-}                                             & \cellcolor[HTML]{EFEFEF}15 pixel (normalized: 0.15) \\ \hline
	\end{tabular}
\end{table}

\begin{figure}[!htbp]
	\setcounter{subfigure}{0}
	\centering
	\begin{minipage}[b]{\textwidth}
		\centering
		\includegraphics[width=0.195\textwidth]{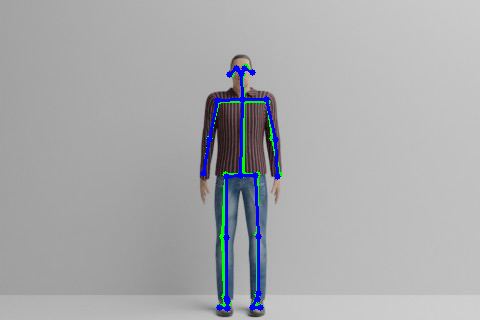}
		\includegraphics[width=0.195\textwidth]{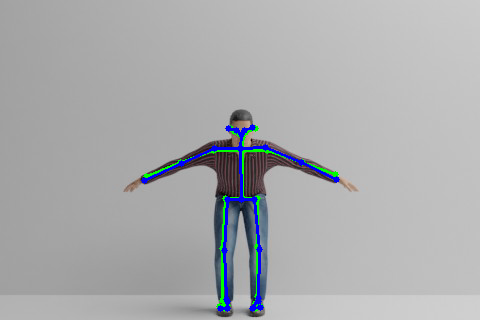}
		\includegraphics[width=0.195\textwidth]{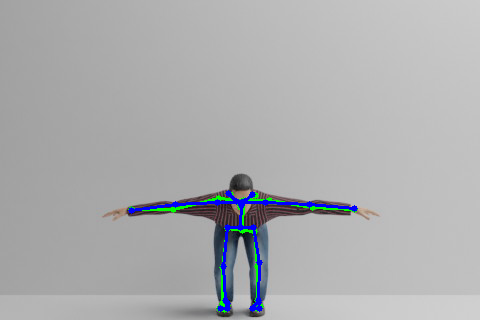}
		\includegraphics[width=0.195\textwidth]{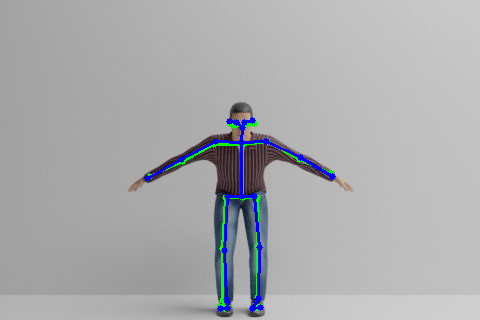}
		\includegraphics[width=0.195\textwidth]{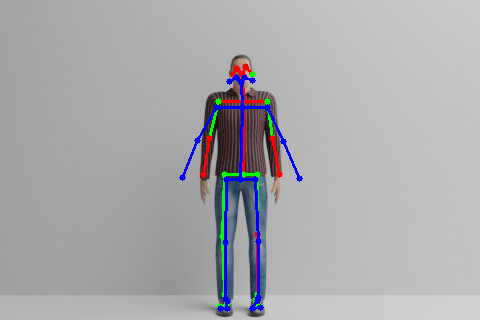}
		\vspace{-0.5cm}
		\subcaption{5 nodes predicted}
		\label{fig:first-experiment-5}
	\end{minipage}
	\vskip\baselineskip
	\begin{minipage}[b]{\textwidth}
		\centering
		\includegraphics[width=0.195\textwidth]{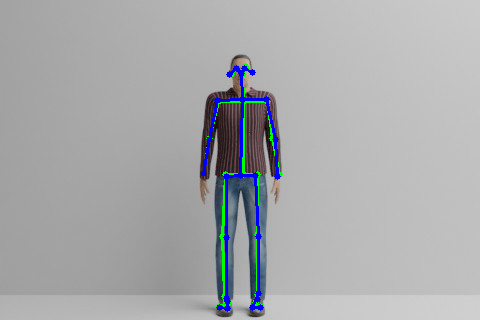}
		\includegraphics[width=0.195\textwidth]{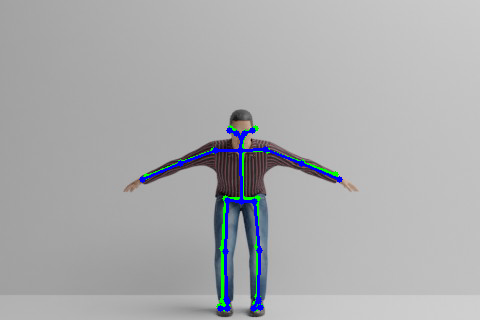}
		\includegraphics[width=0.195\textwidth]{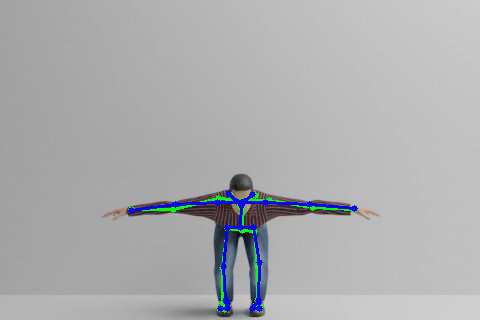}
		\includegraphics[width=0.195\textwidth]{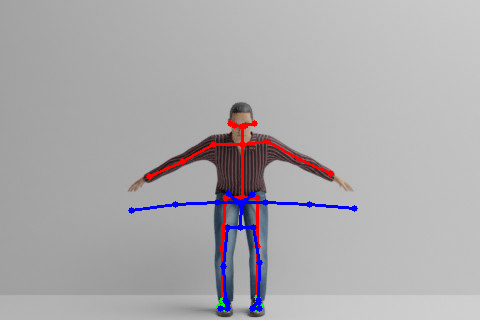}
		\includegraphics[width=0.195\textwidth]{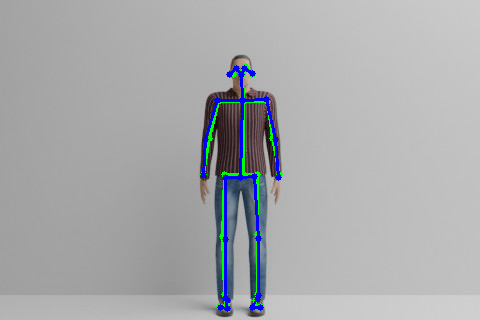}
		\vspace{-0.5cm}
		\subcaption{25 nodes predicted}
		\label{fig:first-experiment-25}
	\end{minipage}
	\vskip\baselineskip
	\begin{minipage}[b]{\textwidth}
		\centering
		\includegraphics[width=0.195\textwidth]{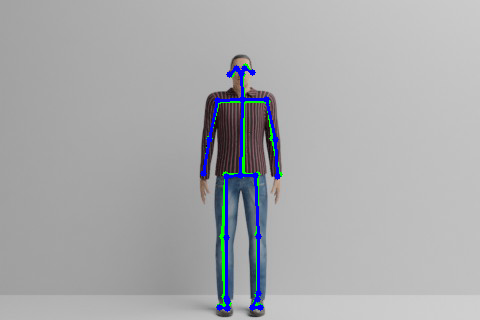}
		\includegraphics[width=0.195\textwidth]{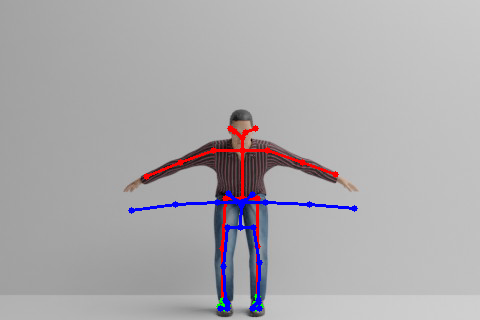}
		\includegraphics[width=0.195\textwidth]{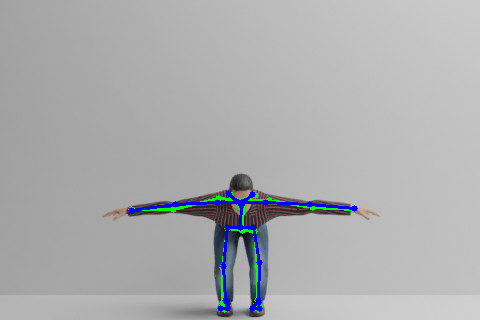}
		\includegraphics[width=0.195\textwidth]{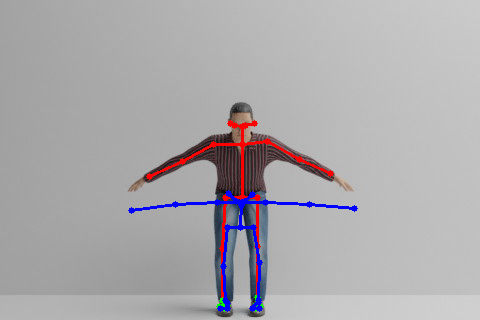}
		\includegraphics[width=0.195\textwidth]{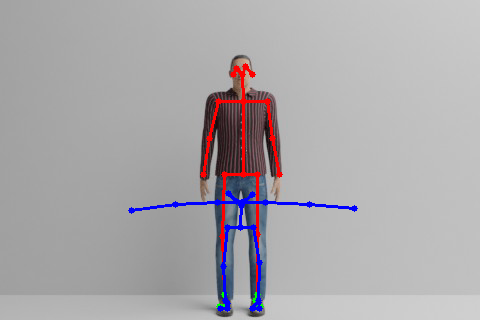}
		\vspace{-0.5cm}
		\subcaption{50 nodes predicted}
		\label{fig:first-experiment-50}
	\end{minipage}
	\caption{Frames 1, 30, 50, 70 and 100 of avatar 01 performing the physical exercise. In green, the real-time human pose estimation of OpenPose is shown and superposed with a blue skeleton that resembles the predicted pose from the Gamma-GWR. Red indicates that the mismatch between prediction and real-time estimation is larger than $d_{t,pose}$ for the given joint.}
	\label{fig:first-experiment}
\end{figure}

The Gamma-GWR predicts the successor node $v$ by creating a merge vector based on the weight and context of the current BMU $u$ comparing it to all node contexts according to
\begin{equation}
	\label{eqn:merge}
	\begin{aligned}
		\mathfrak{s}(u)&=\arg\min_{v\in V}(d_s(u,v)),\\
		d_s(u,v)&=\left\Vert merge(u)-\mathbf{c}_{v,k}\right\Vert_{2}.
	\end{aligned}
\end{equation}
We denote the average joint-wise error over 100 frames per increasing number of predictions in table \ref{tab:first-experiment}.
% If you use beamer only pass "xcolor=table" option, i.e. \documentclass[xcolor=table]{beamer}
\begin{table}[!htbp]
	\centering
	\caption{Average joint-wise error in pixels over 100 frames between key point prediction from Gamma-GWR  (with increasing number of predicted poses up to 100) and OpenPose's real-time estimation. Green indicates the smallest error and red the highest.}
	\label{tab:first-experiment}
	\begin{tabular}{|c|c|c|c|c|c|c|}
		\hline
		\rowcolor[HTML]{C0C0C0} 
		\textbf{Gamma-GWR}                         & \textbf{1}     & \textbf{5}     & \textbf{10} & \textbf{25} & \textbf{50}    & \textbf{100}   \\ \hline
		\cellcolor[HTML]{EFEFEF}\textbf{Nose}      & {\color[HTML]{009901} \textbf{0.349}} & 3.534          & 11.779      & 35.306      & {\color[HTML]{9A0000} \textbf{71.558 }}         & 71.557         \\ \cline{1-1}
		\rowcolor[HTML]{EFEFEF} 
		\textbf{Neck}                              & {\color[HTML]{009901}\textbf{1.292}} & 3.203          & 9.635       & 27.266      & {\color[HTML]{9A0000} \textbf{55.127}}          & 55.125         \\ \cline{1-1}
		\cellcolor[HTML]{EFEFEF}\textbf{RShoulder} & {\color[HTML]{009901}\textbf{0.202}} & 2.645          & 9.118       & 27.505      & {\color[HTML]{9A0000} \textbf{55.958}}          & {\color[HTML]{9A0000} \textbf{55.958}}          \\ \cline{1-1}
		\rowcolor[HTML]{EFEFEF} 
		\textbf{RElbow}                            & {\color[HTML]{009901}\textbf{0.642}} & 1.647          & 6.086       & 19.393      & {\color[HTML]{9A0000} \textbf{39.472}}          & 39.467         \\ \cline{1-1}
		\cellcolor[HTML]{EFEFEF}\textbf{RWrist}    & {\color[HTML]{009901}\textbf{1.426}} & 3.637          & 9.509       & 17.960      & 36.368         & {\color[HTML]{9A0000} \textbf{36.386}}          \\ \cline{1-1}
		\rowcolor[HTML]{EFEFEF} 
		\textbf{LShoulder}                         & {\color[HTML]{009901}\textbf{1.774}} & 3.384          & 9.587       & 27.660      & 56.118         & {\color[HTML]{9A0000} \textbf{56.135}}          \\ \cline{1-1}
		\cellcolor[HTML]{EFEFEF}\textbf{LElbow}    & {\color[HTML]{009901}\textbf{0.693}} & 2.294          & 6.746       & 20.407      & 40.878         & {\color[HTML]{9A0000} \textbf{40.879 }}         \\ \cline{1-1}
		\rowcolor[HTML]{EFEFEF} 
		\textbf{LWrist}                            & {\color[HTML]{009901}\textbf{1.947}} & 3.974          & 9.968       & 19.540      & 38.841         & {\color[HTML]{9A0000} \textbf{38.842}}          \\ \cline{1-1}
		\cellcolor[HTML]{EFEFEF}\textbf{MidHip}    & {\color[HTML]{009901}\textbf{2.154}} & 3.122          & 5.905       & 15.000      & {\color[HTML]{9A0000} \textbf{28.879}}          & {\color[HTML]{9A0000} \textbf{28.879}}          \\ \cline{1-1}
		\rowcolor[HTML]{EFEFEF} 
		\textbf{RHip}                              & {\color[HTML]{009901}\textbf{1.731}} & 2.826          & 5.740       & 14.861      & 28.752         & {\color[HTML]{9A0000} \textbf{28.753}}          \\ \cline{1-1}
		\cellcolor[HTML]{EFEFEF}\textbf{RKnee}     & {\color[HTML]{009901}\textbf{2.263}} & 2.701          & 3.873       & 8.557       & {\color[HTML]{9A0000} \textbf{15.784}}          & 15.771         \\ \cline{1-1}
		\rowcolor[HTML]{EFEFEF} 
		\textbf{RAnkle}                            & 2.095          & {\color[HTML]{009901}\textbf{2.058}} & 2.060       & 1.960       & {\color[HTML]{9A0000} \textbf{1.867 }}          & {\color[HTML]{9A0000} \textbf{1.867}}           \\ \cline{1-1}
		\cellcolor[HTML]{EFEFEF}\textbf{LHip}      & {\color[HTML]{009901}\textbf{1.908}} & 2.644          & 5.451       & 14.485      & 28.466         & {\color[HTML]{9A0000} \textbf{28.467}}          \\ \cline{1-1}
		\rowcolor[HTML]{EFEFEF} 
		\textbf{LKnee}                             & {\color[HTML]{009901}\textbf{1.972}} & 2.455          & 3.944       & 7.908       & {\color[HTML]{9A0000} \textbf{14.338}}          & 14.300         \\ \cline{1-1}
		\cellcolor[HTML]{EFEFEF}\textbf{LAnkle}    & {\color[HTML]{009901}\textbf{1.427}} & 1.509          & 1.830       & 2.119       & {\color[HTML]{9A0000} \textbf{3.018}}           & 3.015          \\ \cline{1-1}
		\rowcolor[HTML]{EFEFEF} 
		\textbf{REye}                              & {\color[HTML]{009901}\textbf{1.311}} & 4.035          & 12.249      & 36.039      & 72.936         & {\color[HTML]{9A0000} \textbf{72.945 }}         \\ \cline{1-1}
		\cellcolor[HTML]{EFEFEF}\textbf{LEye}      & {\color[HTML]{009901}\textbf{1.081}} & 4.111          & 12.396      & 36.011      & {\color[HTML]{9A0000} \textbf{72.610 }}         & 72.600         \\ \cline{1-1}
		\rowcolor[HTML]{EFEFEF} 
		\textbf{REar}                              & {\color[HTML]{009901}\textbf{2.338}} & 3.977          & 10.880      & 32.634      & {\color[HTML]{9A0000} \textbf{65.945}}          & 65.939         \\ \cline{1-1}
		\cellcolor[HTML]{EFEFEF}\textbf{LEar}      & {\color[HTML]{009901}\textbf{1.859}} & 3.491          & 10.571      & 32.553      & {\color[HTML]{9A0000} \textbf{66.190}}          & {\color[HTML]{9A0000} \textbf{66.190}}          \\ \cline{1-1}
		\rowcolor[HTML]{EFEFEF} 
		\textbf{LBigToe}                           & {\color[HTML]{009901}\textbf{3.689}} & 3.699          & 3.691       & 3.735       & 3.817          & {\color[HTML]{9A0000} \textbf{3.818}}           \\ \cline{1-1}
		\cellcolor[HTML]{EFEFEF}\textbf{LSmallToe} & {\color[HTML]{009901}\textbf{0.986}} & 1.004          & 1.094       & 1.345       & 1.740          & {\color[HTML]{9A0000} \textbf{1.744}}           \\ \cline{1-1}
		\rowcolor[HTML]{EFEFEF} 
		\textbf{LHeel}                             & {\color[HTML]{9A0000} \textbf{3.463}}           & 3.375          & 3.330       & 3.092       & 2.812          & {\color[HTML]{009901}\textbf{2.810}} \\ \cline{1-1}
		\cellcolor[HTML]{EFEFEF}\textbf{RBigToe}   & {\color[HTML]{9A0000} \textbf{3.041}}           & 3.033          & 3.019       & 2.881       & {\color[HTML]{009901}\textbf{2.740}} & 2.743          \\ \cline{1-1}
		\rowcolor[HTML]{EFEFEF} 
		\textbf{RSmallToe}                         & {\color[HTML]{009901}\textbf{1.035}} & 1.058          & 1.147       & 1.055       & {\color[HTML]{9A0000} \textbf{1.115}}           & 1.111          \\ \cline{1-1}
		\cellcolor[HTML]{EFEFEF}\textbf{RHeel}     & {\color[HTML]{9A0000} \textbf{2.738}}           & 2.701          & 2.610       & 2.374       & {\color[HTML]{009901}\textbf{1.971}} & 1.996          \\ \hline
		\rowcolor[HTML]{EFEFEF} 
		\rowcolor[HTML]{C0C0C0}\textbf{Average}   & {\color[HTML]{009901}\textbf{1.737}} & 2.885          & 6.489       & 16.466      & {\color[HTML]{9A0000} \textbf{32.292}}          & {\color[HTML]{9A0000} \textbf{32.292 }}         \\ \hline
	\end{tabular}
\end{table}
One can see that for the lower-body joints, especially the feet, the error over all predictions is nearly constant. This is reasonable, since squat exercise does not involve the motion of feet, which results in the constant error for the left and right toes. However, for the upper-body joints and face features, the error increases significantly until 50 predictions. After that, it is nearly identical to the results for 100 predictions. To further understand the error for the upper body, figure \ref{fig:first-experiment} has to be analyzed.
From figure \ref{fig:first-experiment-50} one can see that the architecture is able to process the downward motion, but gets stuck in the first halting point and does not recall the upward motion correctly. Therefore, we assume that the Gamma-GWR gets stuck in a loop of a self-referencing node and thus, cannot predict the upward motion. This also explains the similarity between 50 and 100 predictions, since in both cases, the predictions halts at the same stage of the physical exercise. As a consequence, the question arises, whether a mechanism for recalling a trajectory of BMUs as in the Subnode-GWR and the Episodic-GWR, which does not employ a prediction scheme based on computation but rather on a look-up table, performs better.
\subsection{Comparison Between GWR Variants}
\begin{figure}[!htbp]
	\setcounter{subfigure}{0}
	\centering
	\begin{minipage}[b]{\textwidth}
		\centering
		\includegraphics[width=0.195\textwidth]{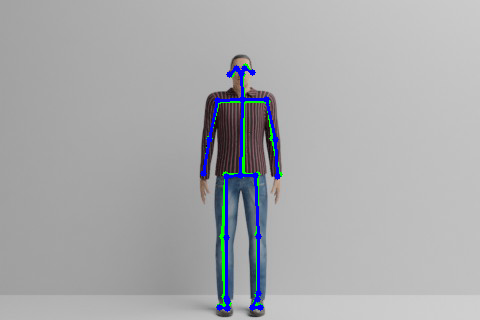}
		\includegraphics[width=0.195\textwidth]{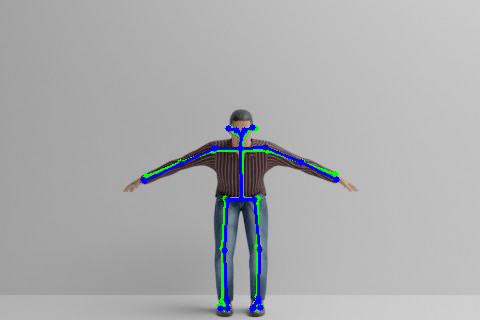}
		\includegraphics[width=0.195\textwidth]{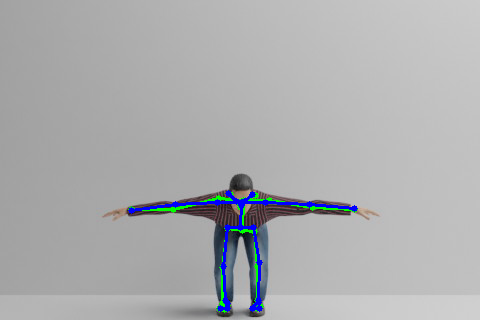}
		\includegraphics[width=0.195\textwidth]{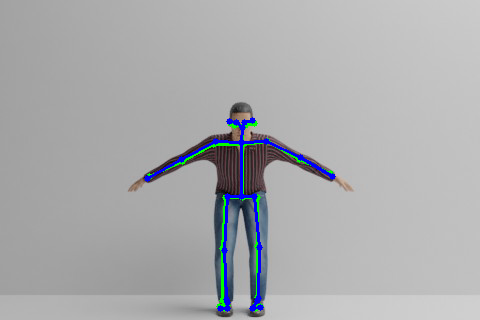}
		\includegraphics[width=0.195\textwidth]{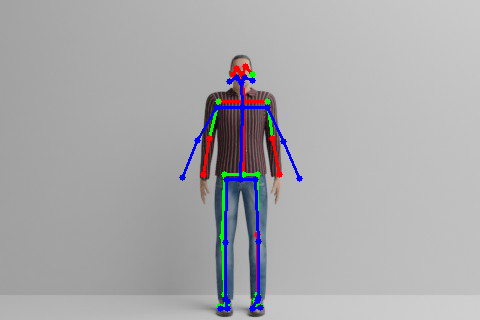}
		\vspace{-0.5cm}
		\subcaption{Gamma-GWR}
		\label{fig:second-experiment-ggwr}
	\end{minipage}
	\vskip\baselineskip
	\begin{minipage}[b]{\textwidth}
		\centering
		\includegraphics[width=0.195\textwidth]{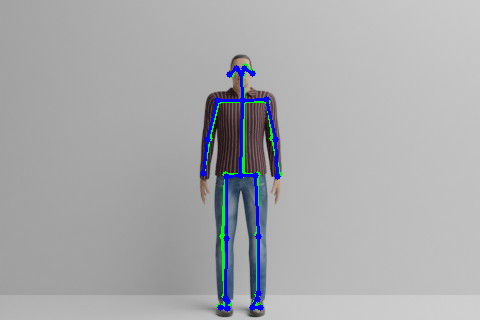}
		\includegraphics[width=0.195\textwidth]{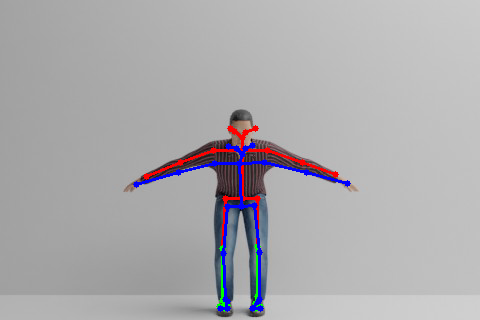}
		\includegraphics[width=0.195\textwidth]{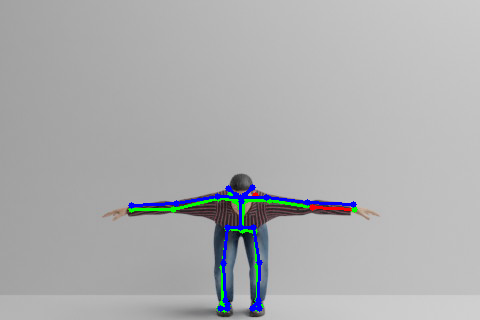}
		\includegraphics[width=0.195\textwidth]{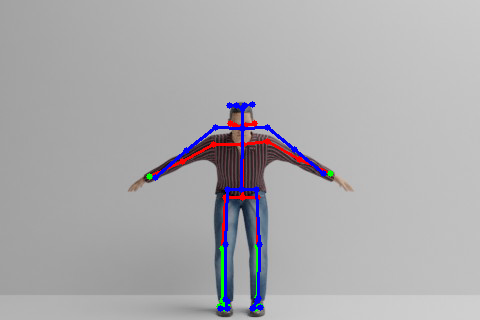}
		\includegraphics[width=0.195\textwidth]{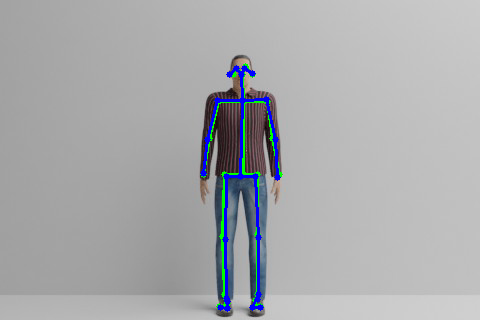}
		\vspace{-0.5cm}
		\subcaption{Episodic-GWR}
		\label{fig:second-experiment-egwr}
	\end{minipage}
	\vskip\baselineskip
	\begin{minipage}[b]{\textwidth}
		\centering
		\includegraphics[width=0.195\textwidth]{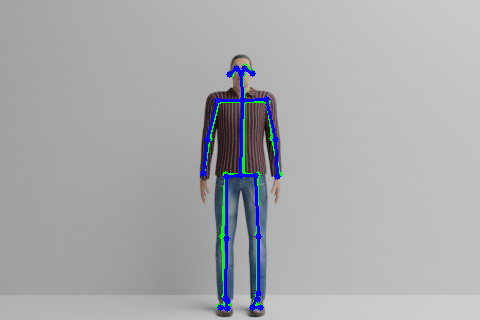}
		\includegraphics[width=0.195\textwidth]{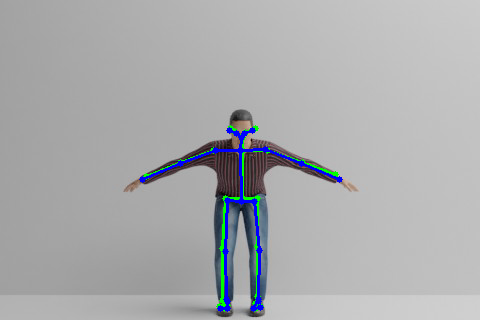}
		\includegraphics[width=0.195\textwidth]{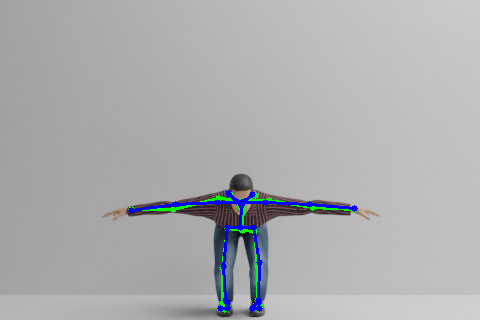}
		\includegraphics[width=0.195\textwidth]{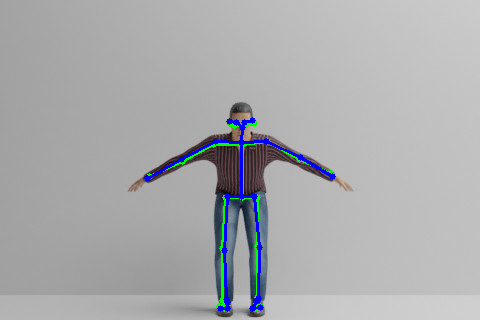}
		\includegraphics[width=0.195\textwidth]{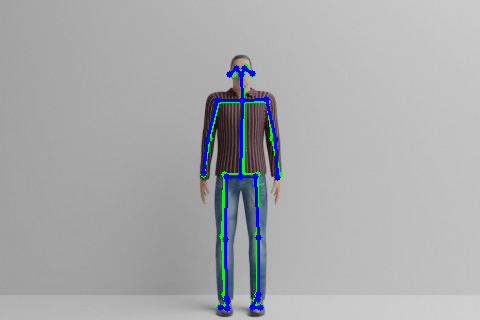}
		\vspace{-0.5cm}
		\subcaption{Subnode-GWR}
		\label{fig:second-experiment-sgwr}
	\end{minipage}
	\caption{Frames 1, 30, 50, 70 and 100 of avatar 01 performing the physical exercise. In green, the real-time human pose estimation of OpenPose is shown and superposed with a blue skeleton that resembles the predicted pose for each architecture. Red indicates that the mismatch between prediction and real-time estimation is larger than $d_{t,pose}$ for the given joint.}
	\label{fig:second-experiment}
\end{figure}

Therefore, in our second experiment, we compare the performance of the Gamma-GWR with 5 predictions against the Episodic-GWR and our proposed architecture the Subnode-GWR. For this experiment, we report in table \ref{tab:second-experiment} the average error for all 25 key points over 100 frames between the real-time estimation of OpenPose of the physical exercise performed by the virtual avatar 01 (see upper left image in figure \ref{fig:avatar-comparison}) and the individual prediction method of each architecture.
\begin{table}[!htbp]
	\centering
	\caption{Average joint-wise error in pixels over 100 frames between key point prediction from Gamma-GWR with 5 predictions, Episodic-GWR as well as Subnode-GWR and OpenPose's real-time estimation. Green indicates the smallest error and red the highest.}
	\label{tab:second-experiment}
	\begin{tabular}{|c|c|c|c|}
		\hline
		\rowcolor[HTML]{C0C0C0} 
		\textbf{Avatar 01}                         & \textbf{Gamma-GWR} & \textbf{Episodic-GWR} & \textbf{Subnode-GWR} \\ \hline
		\cellcolor[HTML]{EFEFEF}\textbf{Nose}      & {\color[HTML]{9A0000} \textbf{3.534}}               & 1.802                 & {\color[HTML]{009901}\textbf{0.402}}       \\ \cline{1-1}
		\rowcolor[HTML]{EFEFEF} 
		\textbf{Neck}                              & {\color[HTML]{9A0000} \textbf{3.203 }}              & 1.631                 & {\color[HTML]{009901}\textbf{1.186}}       \\ \cline{1-1}
		\cellcolor[HTML]{EFEFEF}\textbf{RShoulder} & {\color[HTML]{9A0000} \textbf{2.645}}               & 1.542                 & {\color[HTML]{009901}\textbf{0.334}}       \\ \cline{1-1}
		\rowcolor[HTML]{EFEFEF} 
		\textbf{RElbow}                            & {\color[HTML]{9A0000} \textbf{1.647}}               & 1.546                 & {\color[HTML]{009901}\textbf{0.706}}       \\ \cline{1-1}
		\cellcolor[HTML]{EFEFEF}\textbf{RWrist}    & {\color[HTML]{9A0000} \textbf{3.637 }}              & {\color[HTML]{009901}\textbf{1.337}}        & 1.450                \\ \cline{1-1}
		\rowcolor[HTML]{EFEFEF} 
		\textbf{LShoulder}                         & {\color[HTML]{9A0000} \textbf{3.384 }}              & 2.195                 & {\color[HTML]{009901}\textbf{1.728}}       \\ \cline{1-1}
		\cellcolor[HTML]{EFEFEF}\textbf{LElbow}    & {\color[HTML]{9A0000} \textbf{2.294 }}              & 1.021                 & {\color[HTML]{009901}\textbf{0.611}}       \\ \cline{1-1}
		\rowcolor[HTML]{EFEFEF} 
		\textbf{LWrist}                            & {\color[HTML]{9A0000} \textbf{3.974}}               & {\color[HTML]{009901}\textbf{1.691}}        & 1.868                \\ \cline{1-1}
		\cellcolor[HTML]{EFEFEF}\textbf{MidHip}    & {\color[HTML]{9A0000} \textbf{3.122}}               & {\color[HTML]{009901}\textbf{1.884}}        & 2.124                \\ \cline{1-1}
		\rowcolor[HTML]{EFEFEF} 
		\textbf{RHip}                              & {\color[HTML]{9A0000} \textbf{2.826 }}              & {\color[HTML]{009901}\textbf{1.338}}        & 1.676                \\ \cline{1-1}
		\cellcolor[HTML]{EFEFEF}\textbf{RKnee}     & {\color[HTML]{9A0000} \textbf{2.701}}               & {\color[HTML]{009901}\textbf{2.120}}        & 2.308                \\ \cline{1-1}
		\rowcolor[HTML]{EFEFEF} 
		\textbf{RAnkle}                            & {\color[HTML]{009901}\textbf{2.058}}              & {\color[HTML]{9A0000} \textbf{2.119 }}                 & 2.105       \\ \cline{1-1}
		\cellcolor[HTML]{EFEFEF}\textbf{LHip}      & {\color[HTML]{9A0000} \textbf{2.644 }}              & {\color[HTML]{009901}\textbf{1.775}}        & 1.846                \\ \cline{1-1}
		\rowcolor[HTML]{EFEFEF} 
		\textbf{LKnee}                             & {\color[HTML]{9A0000} \textbf{2.455 }}              & {\color[HTML]{009901}\textbf{1.754}}        & 1.977                \\ \cline{1-1}
		\cellcolor[HTML]{EFEFEF}\textbf{LAnkle}    & {\color[HTML]{9A0000} \textbf{1.509}}               & {\color[HTML]{009901}\textbf{1.415}}        & 1.461                \\ \cline{1-1}
		\rowcolor[HTML]{EFEFEF} 
		\textbf{REye}                              & {\color[HTML]{9A0000} \textbf{4.035}}               & 1.999                 & {\color[HTML]{009901}\textbf{1.211}}       \\ \cline{1-1}
		\cellcolor[HTML]{EFEFEF}\textbf{LEye}      & {\color[HTML]{9A0000} \textbf{4.111}}               & 1.582                 & {\color[HTML]{009901}\textbf{0.846}}       \\ \cline{1-1}
		\rowcolor[HTML]{EFEFEF} 
		\textbf{REar}                              & {\color[HTML]{9A0000} \textbf{3.977}}               & 2.812                 & {\color[HTML]{009901}\textbf{2.319}}       \\ \cline{1-1}
		\cellcolor[HTML]{EFEFEF}\textbf{LEar}      & {\color[HTML]{9A0000} \textbf{3.491}}               & 2.716                 & {\color[HTML]{009901}\textbf{1.936}}       \\ \cline{1-1}
		\rowcolor[HTML]{EFEFEF} 
		\textbf{LBigToe}                           & {\color[HTML]{009901}\textbf{3.699}}              & 3.707        & {\color[HTML]{9A0000} \textbf{3.708}}                 \\ \cline{1-1}
		\cellcolor[HTML]{EFEFEF}\textbf{LSmallToe} & {\color[HTML]{9A0000} \textbf{1.004}}               & {\color[HTML]{009901}\textbf{0.966}}        & 0.976                \\ \cline{1-1}
		\rowcolor[HTML]{EFEFEF} 
		\textbf{LHeel}                             & {\color[HTML]{009901}\textbf{3.375}}     & {\color[HTML]{9A0000} \textbf{3.523 }}                 & 3.481                \\ \cline{1-1}
		\cellcolor[HTML]{EFEFEF}\textbf{RBigToe}   & {\color[HTML]{009901}\textbf{3.033}}     & {\color[HTML]{9A0000} \textbf{3.054}}                  & 3.049                \\ \cline{1-1}
		\rowcolor[HTML]{EFEFEF} 
		\textbf{RSmallToe}                         & {\color[HTML]{9A0000} \textbf{1.058}}               & {\color[HTML]{009901}\textbf{1.045}}        & 1.054                \\ \cline{1-1}
		\cellcolor[HTML]{EFEFEF}\textbf{RHeel}     & {\color[HTML]{009901}\textbf{2.701}}     & {\color[HTML]{9A0000} \textbf{2.809}}                  & 2.799                \\ \hline
		\rowcolor[HTML]{C0C0C0} 
		\textbf{Average}                           & {\color[HTML]{9A0000} \textbf{2.885}}               & 1.975                 & {\color[HTML]{009901}\textbf{1.726}}       \\ \hline
	\end{tabular}
\end{table}
With an average error of 1.726, the Subnode-GWR performs best, with the Episodic-GWR ranking second with 1.975, leaving the Gamma-GWR behind with 2.885. The results show that the prediction algorithm of the Gamma-GWR lacks behind the approach of the Episodic-GWR. In figure \ref{fig:second-experiment}, however, it becomes obvious that disallowing nodes to reference to themselves leads to asynchronous predictions. Nodes missing in $P$ is equivalent to skipping frames in the rendered video. Therefore, the predicted blue skeleton performs the exercise faster than the virtual avatar. This issue is overcome by the Subnode-GWR, which triggers no erroneous feedback according to figure \ref{fig:second-experiment-sgwr}, distinguishing the Subnode-GWR as the best approach for the task at hand. Though, while it performs well on virtual avatar 01, on which it has been trained on, the Subnode-GWRs novelty lies in its continual learning scheme, which is evaluated in a third experiment.
\subsection{Continual Learning of Subnode-GWR}
To test that the Subnode-GWR is able to learn continuously, we train the network on the virtual avatar 01 and then online on the remaining 9 virtual avatars. Therefore, we classify whether each joint has been correctly marked as erroneous or not in comparison with OpenPose's real-time estimation and the embedded poses in the subnodes. Therefore, we defined a ground truth for every exercise and its common errors, that indicates which joint should be drawn in red. The results of the classification are shown in table \ref{tab:third-experiment}.
\begin{table}[!htbp]
	\centering
	\caption{Accuracy and standard deviation for classifying joints over all avatars performing the exercise including common errors correctly as right or wrong based on $d_{t,pose}$ for a centered position in the field of view of the camera (no rotation or translation).}
	\label{tab:third-experiment}
	\begin{tabular}{|c|c|c|c|l|l|l|}
		\hline
		\rowcolor[HTML]{C0C0C0} 
		\textbf{Centered}                   & \textbf{Correct} & \textbf{Arms} & \textbf{Head} & \textbf{Legs}                                      & \textbf{Side}                                                             & \textbf{Speed}                                     \\ \hline
		\cellcolor[HTML]{EFEFEF}\textbf{Avatar 01} & 1.000            & 1.000         & 1.000         & 1.000                                              & \multicolumn{1}{l|}{1.000}                                                & 1.000                                              \\ \cline{1-1}
		\rowcolor[HTML]{EFEFEF} 
		\textbf{Avatar 02}                         & 1.000            & 1.000         & 0.880         & {\color[HTML]{333333} 1.000}                       & \multicolumn{1}{l|}{\cellcolor[HTML]{EFEFEF}{\color[HTML]{333333} 0.840}} & {\color[HTML]{333333} 1.000}                       \\ \cline{1-1}
		\cellcolor[HTML]{EFEFEF}\textbf{Avatar 03} & 1.000            & 0.760         & 0.920         & 1.000                                              & \multicolumn{1}{l|}{0.800}                                                & 1.000                                              \\ \cline{1-1}
		\rowcolor[HTML]{EFEFEF} 
		\textbf{Avatar 04}                         & 1.000            & 0.920         & 0.480         & 1.000                                              & \multicolumn{1}{l|}{\cellcolor[HTML]{EFEFEF}1.000}                        & 1.000                                              \\ \cline{1-1}
		\cellcolor[HTML]{EFEFEF}\textbf{Avatar 05} & 1.000            & 0.520         & 0.520         & 0.960                                              & \multicolumn{1}{l|}{0.880}                                                & 1.000                                              \\ \cline{1-1}
		\rowcolor[HTML]{EFEFEF} 
		\textbf{Avatar 06}                         & 1.000            & 1.000         & 0.600         & 1.000                                              & \multicolumn{1}{l|}{\cellcolor[HTML]{EFEFEF}1.000}                        & 1.000                                              \\ \cline{1-1}
		\cellcolor[HTML]{EFEFEF}\textbf{Avatar 07} & 1.000            & 1.000         & 0.920         & 1.000                                              & \multicolumn{1}{l|}{0.960}                                                & 1.000                                              \\ \cline{1-1}
		\rowcolor[HTML]{EFEFEF} 
		\textbf{Avatar 08}                         & 1.000            & 0.960         & 1.000         & 1.000                                              & \multicolumn{1}{l|}{\cellcolor[HTML]{EFEFEF}1.000}                        & 1.000                                              \\ \cline{1-1}
		\cellcolor[HTML]{EFEFEF}\textbf{Avatar 09} & 1.000            & 1.000         & 0.360         & 1.000                                              & \multicolumn{1}{l|}{0.880}                                                & 0.080                                              \\ \cline{1-1}
		\rowcolor[HTML]{EFEFEF} 
		\textbf{Avatar 10}                         & 1.000            & 0.320         & 0.480         & 1.000                                              & \multicolumn{1}{l|}{\cellcolor[HTML]{EFEFEF}0.920}                        & 1.000                                              \\ \hline
		\rowcolor[HTML]{C0C0C0} 
		\textbf{Average}                    & 1.00             & 0.848         & 0.716         & \multicolumn{1}{c|}{\cellcolor[HTML]{C0C0C0}0.996} & \multicolumn{1}{c|}{\cellcolor[HTML]{C0C0C0}0.928}                        & \multicolumn{1}{c|}{\cellcolor[HTML]{C0C0C0}0.908} \\ \hline
		\rowcolor[HTML]{C0C0C0} 
		\textbf{Std. Dev.}                  & 0.00             & 0.242         & 0.250         & \multicolumn{1}{c|}{\cellcolor[HTML]{C0C0C0}0.013} & \multicolumn{1}{c|}{\cellcolor[HTML]{C0C0C0}0.075}                        & \multicolumn{1}{c|}{\cellcolor[HTML]{C0C0C0}0.291} \\ \hline
	\end{tabular}
\end{table}
For the correct performance of the exercise, the Subnode-GWR is able to give accurate feedback for all joints. But, we can see, that the accuracy reduces to $71,6\%$ for exercise \ref{fig:avatars-exercises-head}. One can see that for avatar 09 (see figure \ref{fig:avatar-comparison}), the accuracy is significantly lower in comparison to other avatars and common errors. This repeats for the common error in figure \ref{fig:avatars-exercises-speed}, where the user performs the exercise too fast. After further investigation, we conclude that inaccuracy results from the Subnode-GWR selecting the wrong subnode at the first frame.

This is due to the fact, that avatar 09 resembles many other avatars with nearly matching height and weight features. This leads to the question, how robust the approach overall is against variations in e. g. rotation and translation.
\subsection{Robustness Against Rotation and Translation}
Thus, in order to further evaluate the robustness and spot possible drawbacks of the approach, we conduct a final fourth experiment, where we rotate each avatar for every exercise by 5 degrees, translate them by 5 cm to the left and lastly combine both rotation and translation.
\begin{table}[!htbp]
	\centering
	\caption{Average accuracy and standard deviation for classifying joints over exercise including common errors with deferring positions (rotation: 5 degrees, translation: 5cm) correctly as right or wrong based on $d_{t,pose}$.}
	\label{tab:fourth-experiment}
	\begin{tabular}{|c|c|c|c|c|}
		\hline
		\rowcolor[HTML]{C0C0C0} 
		\textbf{Variation}                       & \textbf{Centered} & \textbf{Rotation} & \textbf{Translation} & \multicolumn{1}{l|}{\cellcolor[HTML]{C0C0C0}\textbf{Rot. + Trans.}} \\ \hline
		\cellcolor[HTML]{EFEFEF}\textbf{Correct} & 1.000             & 0.724             & 0.980                & 0.812                                                               \\ \cline{1-1}
		\rowcolor[HTML]{EFEFEF} 
		\textbf{Arms}                            & 0.848             & 0.880             & 0.860                & {\color[HTML]{333333} 0.720}                                        \\ \cline{1-1}
		\cellcolor[HTML]{EFEFEF}\textbf{Head}    & 0.716             & 0.876             & 0.752                & 0.812                                                               \\ \cline{1-1}
		\rowcolor[HTML]{EFEFEF} 
		\textbf{Legs}                            & 0.996             & 0.996             & 0.996                & 0.988                                                               \\ \cline{1-1}
		\cellcolor[HTML]{EFEFEF}\textbf{Side}    & 0.928             & 0.932             & 0.920                & 0.772                                                               \\ \cline{1-1}
		\rowcolor[HTML]{EFEFEF} 
		\textbf{Speed}                           & 0.908             & 0.908             & 0.900                & 0.904                                                               \\ \hline
		\rowcolor[HTML]{C0C0C0} 
		\textbf{Average}                         & 0.899             & 0.886             & 0.901                & 0.835                                                               \\ \hline
		\rowcolor[HTML]{C0C0C0} 
		\textbf{Std. Dev.}                       & 0.106             & 0.091             & 0.089                & 0.096                                                               \\ \hline
	\end{tabular}
\end{table}
In table \ref{tab:fourth-experiment}, the accuracy for a centered view without rotation and translation is $89,9\%$, with rotation $88,6\%$, with translation $90,1\%$ and finally with both rotation and translation combined $83,5\%$. Overall, the network seems to be unaffected by translation, leading to a small increase in accuracy surprisingly. Rotating the avatars by 5 degrees leads to a small drop in accuracy of around $1\%$. Combining both rotation and translation reduces the accuracy by around $6\%$. Most influential to this drop is the common error, where the upper body is tilted with an accuracy of $77,2\%$. Still, reflecting on all common errors in execution and keeping in mind that they have been exaggerated for the avatars in order to properly evaluate robustness on challenging tasks, we feel that with an average accuracy of $88\%$ over all variations and disturbances the Subnode-GWR to be robust against perturbations commonly occurring in the application.
\section{Conclusion and Future Work}\label{conclusion}
Physical exercise is a precondition for a healthy lifestyle, but requires proper technique in order to prevent injuries. To support on this matter, we  employed the humanoid robot Pepper as a motivator and feedback giver and developed the GWR algorithm with subnodes and a continual learning scheme, which we call Subnode-GWR. While the proposed architecture works well within its purpose, there are still caveats that can be improved. For one, the Subnode-GWR tackles catastrophic forgetting by increasing the capacity of the network rather than restructuring knowledge. This is of course a drawback of the Grow-When-Required approach itself, which has not been solved yet and requires future work. Secondly, the Subnode-GWR requires carefully monitored input from a supervisor (e. g. a physiotherapist) during the learning phase, since its adaptivity is limited within a range of tolerance, that has to be tuned manually. Here, future work could improve on the adaption process, making it self-sustained, not requiring for additional supervision. Still, we evaluated the Subnode-GWR against already existing GWR variants (Gamma-GWR and Episodic-GWR) and showed the advantages of it. We also examined in further experiments the capabilities of the Subnode-GWR regarding continual learning on multiple avatars and the robustness  against rotation and translation. To do so, we recorded the Virtual-Squat dataset with 10 virtual avatars that allow for an evaluation without human participants. Finally, we also see the usage of the Subnode-GWR not limited to its current application, but it can rather be beneficial in cases where a  robust and precise replay of information, e. g., as an episodic memory, is required.
\section*{Conflict of Interest Statement}
%All financial, commercial or other relationships that might be perceived by the academic community as representing a potential conflict of interest must be disclosed. If no such relationship exists, authors will be asked to confirm the following statement: 
The authors declare that the research was conducted in the absence of any commercial or financial relationships that could be construed as a potential conflict of interest.
\section*{Author Contributions}
ND and MK conceived the presented experiments. ND developed the neural architectures and conducted and evaluated the experiments with support from MK. MK created the dataset. ND was the primary contributor to the final version of the manuscript. SW and MK planned the research study, proposal, and supervised the project and revised the manuscript. All authors provided critical feedback and helped shape the research, analysis and manuscript.
\section*{Funding}
This research was partially supported by the Federal Ministry for Economic Affairs and Energy (BMWi) under the project KI-SIGS and the German Research Foundation (DFG) under the project Transregio Crossmodal Learning (TRR 169). 
%
%\section*{Acknowledgments}
%This is a short text to acknowledge the contributions of specific colleagues, institutions, or agencies that aided the efforts of the authors.
%
%\section*{Data Availability Statement}
%The full Virtual-Squat dataset will be made available at: \url{https://www.inf.uni-hamburg.de/en/inst/ab/wtm/research/corpora.html}
% Please see the availability of data guidelines for more information, at https://www.frontiersin.org/about/author-guidelines#AvailabilityofData
\bibliographystyle{frontiersinSCNS_ENG_HUMS} % for Science, Engineering and Humanities and Social Sciences articles, for Humanities and Social Sciences articles please include page numbers in the in-text citations

\end{document}